\documentclass[conference]{IEEEtran}

\usepackage{cite}
\usepackage{amsmath,amssymb,amsfonts}
\usepackage{algorithmic}
\usepackage{algorithm} 
\usepackage{graphicx}
\usepackage{url}
\usepackage{textcomp}
\usepackage{hyperref}
\usepackage[table,xcdraw]{xcolor}
\usepackage{xcolor}
\usepackage{tikz}
\usepackage{enumitem}

\usepackage{multirow}

\usepackage{microtype}

\def\BibTeX{{\rm B\kern-.05em{\sc i\kern-.025em b}\kern-.08em
    T\kern-.1667em\lower.7ex\hbox{E}\kern-.125emX}}
\usepackage{todonotes}

\definecolor{mygreen}{RGB}{14, 107, 14}
\definecolor{myred}{RGB}{151, 0, 0}

\newcommand{\greencheck}{}%
\DeclareRobustCommand{\greencheck}{%
  \tikz\fill[scale=0.4, color=mygreen]
  (0,.35) -- (.25,0) -- (1,.7) -- (.25,.15) -- cycle;%
}

\newcommand{\crosscheck}{$\mathbin{\tikz [x=2.4ex,y=2.4ex,line width=.4ex, myred, scale=0.5] \draw (0,0) -- (1,1) (0,1) -- (1,0);}$}%

\begin{document}

\title{DiCE4EL: Interpreting Process Predictions using a Milestone-Aware Counterfactual Approach}

\author{\IEEEauthorblockN{Chihcheng Hsieh}
\IEEEauthorblockA{\textit{Queensland University of Technology} \\
Brisbane, Australia \\
hsiehc2@qut.edu.au}
\and
\IEEEauthorblockN{Catarina Moreira}
\IEEEauthorblockA{\textit{Queensland University of Technology} \\
Brisbane, Australia \\
catarina.pintomoreira@qut.edu.au}
\and
\IEEEauthorblockN{Chun Ouyang}
\IEEEauthorblockA{\textit{Queensland University of Technology} \\
Brisbane, Australia \\
c.ouyang@qut.edu.au}

\and
\IEEEauthorblockN{~~~~~~~~~~~~~~~~~~~~~~~~~~~~~~Paper accepted at the } 
\IEEEauthorblockA{~~~~~~~~~~~~~~~~~~~~~~~~~~~~~~~~~\textit{Proceedings of the 3rd International Conference on Process Mining (ICPM2021)}}

}

\maketitle
\begin{abstract}
Predictive process analytics often apply machine learning to predict the future states of a running business~process. However, the internal mechanisms of many existing predictive algorithms are opaque and a human decision-maker is unable to understand \emph{why} a certain activity was predicted. 
Recently, counterfactuals have been proposed in the literature to derive human-understandable explanations from predictive models. Current counterfactual approaches consist of finding the minimum feature change that can make a certain prediction flip its outcome. Although many algorithms have been proposed, their application to multi-dimensional sequence data like event logs has not been explored in the literature.

In this paper, we explore the use of a recent, popular model-agnostic counterfactual algorithm, DiCE, in the context of predictive process analytics. The analysis reveals that DiCE is unable to derive explanations for process predictions, due to 
(1) process domain knowledge not being taken into account, 
(2) long traces of process execution that often tend to be less understandable, and 
(3) difficulties in optimising the counterfactual search with categorical variables. 
We design an extension of DiCE, namely DiCE4EL (DiCE for Event Logs), that can generate counterfactual explanations for process prediction, and propose an approach that supports deriving milestone-aware counterfactual explanations at key intermediate stages along process execution to promote interpretability. We apply our approach to a publicly available real-life event log and the analysis results demonstrate the effectiveness of the proposed approach.
\end{abstract}


\section{Introduction} 
\label{sec:intro}


Predictive process analytics often apply machine learning to predict the future states of a running business process~\cite{Teinemaa19,Verenich19}. Especially in recent years, the application of deep learning techniques has received great attention in predictive process analytics~\cite{neu2021systematic}, given their capability in generating accurate predictions. Despite their success, the internal mechanisms of machine/deep learning models are an enigma because humans cannot scrutinize how these intelligent systems operate and what they do. This is known as the \textit{black-box} problem~\cite{Lipton2018}. 

Several factors motivated the need to turn black-box predictions transparent to human decision-makers, and this has given the rise of a new research field called explainable AI (XAI). XAI is the set of methods that allows human users to comprehend and trust the results and output created by machine learning algorithms. This implies describing an AI model, its expected impact, and potential biases~\cite{Doran2017}. 


We propose the topic of explainable predictive process analytics, which refers to approaches, methods, and techniques that aim to provide explanations of how and why a specific trace of process execution obtained a particular prediction. An overarching goal is to generate human-understandable, context-relevant explanations of the reasoning of specific predictions learned by an underlying process predictive model or algorithm. Also, for a model to be interpretable, it must suggest explanations that make sense not only to the decision-maker but also ensure that the predictions are in accordance with the process domain knowledge and that explanations accurately represent the actual reasons for the model's decisions~\cite{serrano2019attention,Moreira20medical}. 

Next event prediction is an important task in predictive process analytics~\cite{EVERMANN2017129}. Applying next event prediction iteratively and progressively makes it possible to obtain a sequence of future events, which corresponds to the prediction of a remaining sequence of a particular trace. This type of analysis can equip business analysts with insights that can help them to target potential unexpected outcomes, predict time-consuming process execution, and thus prepare strategies to mitigate relevant issues and problems in a timely manner. 
Explainability will complement this process by providing justifications of why the underlying predictive model is making certain predictions step-by-step alongside process execution, which will help engage process stakeholders in the prediction process. 

There exist several methods that can be used to achieve explainability (e.g., LIME~\cite{Ribeiro16}, SHAP~\cite{Lundberg17}, LINDA-BN~\cite{Moreira21dss}). In this paper, we explore approaches that have been so far underexplored in the literature of predictive process analytics --- \textit{counterfactuals explanations}.
Counterfactuals are considered a fundamental approach to achieve explainability, responsibility, and accountability in AI systems. Counterfactuals  (from ‘‘contrary-to-fact’’~\cite{Byrne1997}) describe events or states of the world that implicitly or explicitly contradict factual knowledge. In predictive process analytics, counterfactuals are an important input for helping business stakeholders understand from a prediction what underlying activities would have to be changed in order to achieve the desired outcomes.

\subsection{Counterfactuals in XAI}

Counterfactuals are a conditional assertion whose antecedent is false and whose consequent describes how the world would have been if the antecedent had occurred (a \textit{what-if} question). In the field of XAI, counterfactuals provide interpretations as a means to point out \textit{which changes would be necessary to accomplish the desired goal} (prediction), supporting the understanding of why the current situation had a certain predictive outcome. Most XAI approaches tend to focus on answering \textit{why} a black-box predicted a certain outcome. Counterfactuals, on the other hand, attempt to answer this question through a hypothetical \textit{what-if} scenario, which implies knowing \textit{what features would the user need to change to achieve a desired outcome}~\cite{Poyiadzi20}.

In process analytics, counterfactual explanations enable users to query the predictions of autonomous systems by posing a hypothetical scenario 
and questioning what would need to be done to achieve a desired outcome. For instance, consider the scenario where a machine learning algorithm assesses whether a person should be granted a loan or not, and the algorithm predicts that the person had the loan denied. An example of a counterfactual explanation for this outcome would be: \textit{if the person's income were higher than \$15,000, then he/she would have been granted a loan}~\cite{Mothilal20}.

Before we proceed, it is important to clarify the difference between predictive models and counterfactual models. 
A predictive model attempts to foresee the future state from observed data, e.g. to answer the question like ``after I take medication~$X$, will my headache disappear?" In a prediction task, an event we wish to predict did not occur at the moment of the query. 
In a counterfactual approach, the event has already taken place. This implies that all data has already been observed, including the data that resulted from some action or decision. In this case, counterfactual models answer the question like ``I had a headache, and after I took medication~$X$, my headache is now gone. What if I had chosen medication~$Y$, would I still have a headache?''. In this example, one cannot go back in time and change to medication~$Y$ to observe the effect since a decision to take medication~$X$ already took place, and the data has already been collected. Counterfactual approaches can help in these scenarios by providing a hypothetical scenario where such variable changes would have occurred. This approach is highly useful in predictive process analytics, where alternative process execution scenarios could be provided to the decision-maker as counterfactual explanations of why a particular outcome occurred.

\subsection{The Need for Novel Counterfactual Approaches in PPA}

Although counterfactual approaches for XAI are a relatively recent topic, so far, many algorithms have already been~proposed (refer to~\cite{Moreira21infofusion,Verma20} for recent surveys). However, in what concerns predictive process analytics (PPA) and to the best of our knowledge, counterfactual approaches have never been investigated under the realm of explaining process predictive models underpinned by machine intelligence. The nature of event logs used in this domain imposes new challenges in finding counterfactuals for process prediction. For example, existing counterfactual approaches focus on traditional datasets, which often deal with one perspective in identifying a minimum change in features that will allow a predictive model ``to flip'' its prediction and generate the counterfactual scenario. However, when it comes to event logs, existing algorithms do not work due to the multi-perspective nature of process execution recorded by event logs. Consequently, new solutions are needed to support generating counterfactual explanations for process prediction. 



\subsection{Contributions}

We investigate the potential of applying existing counterfactual algorithms in XAI literature to predictive process~analytics. We explore the use of a recent, popular counterfactual algorithm, DiCE~\cite{Mothilal20}, that generates diverse counterfactual explanations. We chose DiCE, because it satisfies several essential properties for generating effective counterfactuals. 
These properties include \textit{plausibility}, \textit{proximity}, \textit{diversity}, and \textit{sparsity}~\cite{Moreira21infofusion}. Our preliminary experiments reveal that DiCE is unable to derive counterfactual explanations using event logs. 

We propose an extension of DiCE, namely DiCE4EL (which stands for DiCE for Event Logs), that aims to support generating counterfactual explanations for process prediction. Our work in this paper focuses on next event prediction for generating counterfactual explanations and propose a milestone-aware approach when analysing counterfactual explanations. The notion of \textit{milestones} is used to capture those intermediate stages that one would expect to reach in order to achieve a desirable final outcome at the end of process execution. The proposed approach is evaluated using a publicly available real-life event log concerning loan applications from a financial institution and widely used in predictive process analytics. 

The main contributions of this paper are as follows:
\begin{itemize}
    \item We propose a generic neural network architecture for predicting next event that takes into account both dynamic and static features;
    \item We propose DiCE4EL, an extension of DiCE algorithm, that can be applied to event logs and generate counterfactual explanations for process prediction; and 
    \item We propose a milestone-aware approach to analyse counterfactual explanations step-by-step alongside process execution thus further improving model interpretability.
\end{itemize}




\section{DiCE: Diverse Counterfactual Explanations}
\label{sec:dice}

DiCE (Diverse Counterfactual Explanations) was initially proposed by Mothilal et al. \cite{Mothilal20}. Given a query feature vector with a prediction $y$, DiCE computes a set of diverse counterfactual explanations by finding candidate feature vectors that are close to the query instance but with an opposite prediction $\bar{y}$. This diversity of explanations allows the user to choose counterfactuals that are more understandable and, consequently, more interpretable.  In DiCE, diversity is formalized as a determinant point process based on the determinant of the matrix containing information about the distances between a counterfactual candidate instance and the query instance to be explained. However, when we apply DiCE to event logs, we usually obtain one of two outcomes: (1) either the algorithm cannot find any counterfactual, or (2) the counterfactuals generated make no sense according to the domain process knowledge. 

\begin{figure}[h!]
    \centering
    \includegraphics[scale=0.12]{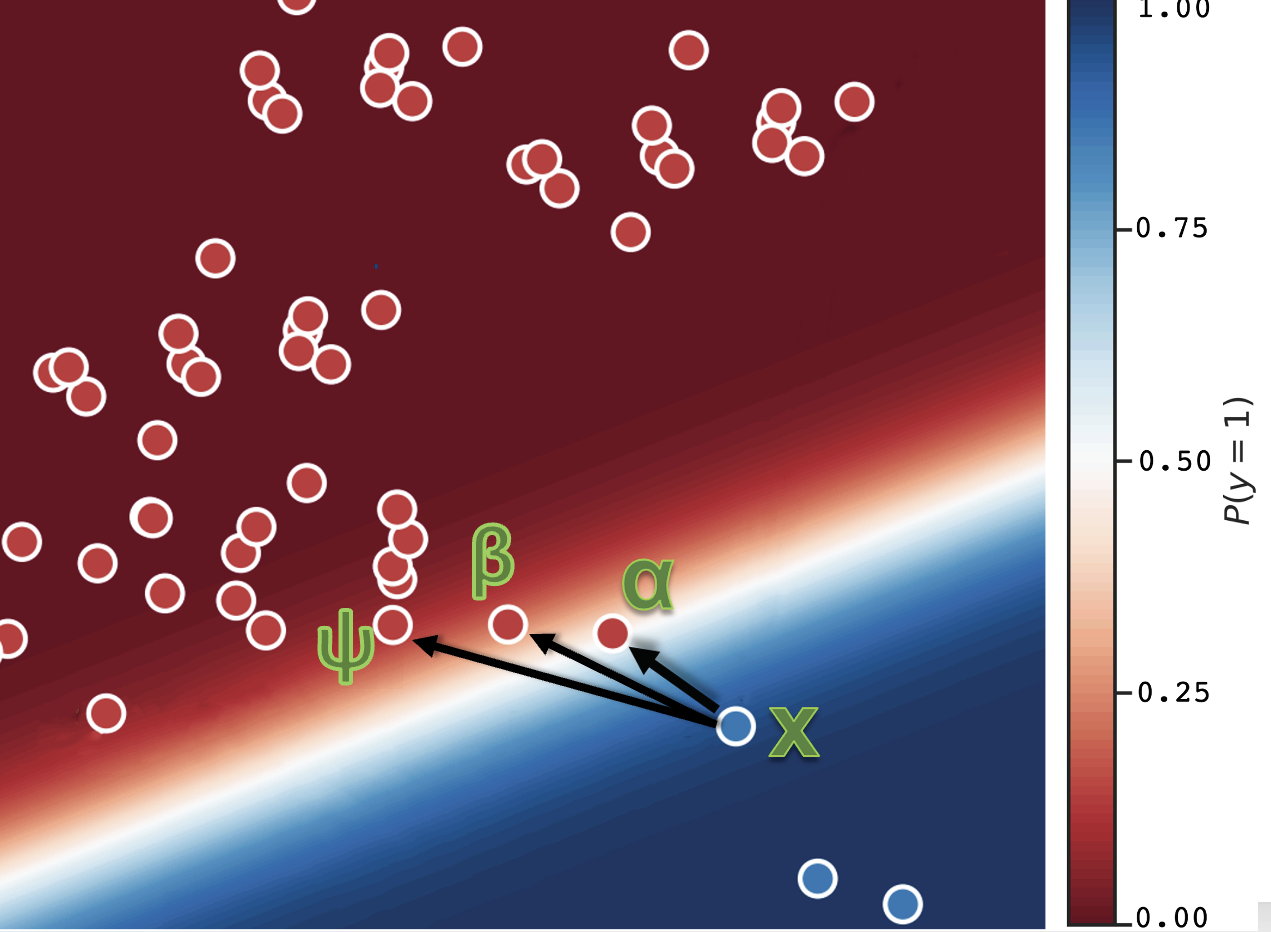}
    \caption{Example of counterfactuals returned by DiCE. Given a datapoint $x$ that we want to explain, the goal is to find the minimum set of features that need to be changed for $x$ to \textit{flip} the prediction. In this figure, $\alpha$ corresponds to the counterfactual that has the minimum distance to to $x$ and $\beta$ and $psi$ correspond to diverse counterfactual possibilities of $x$.}
    \label{fig:counterfactual_general}
\end{figure}

 Figure~\ref{fig:counterfactual_general} shows an illustration of several counterfactual candidates for a data instance $x$ according to DiCE.

In DiCE, the loss function is presented in Equation~\ref{eq:dice}, and is given by a linear combination of three components: (1) a hinge loss function that is a metric that minimizes the distance between the user prediction $f(.)'$ for $c_is$ and an ideal outcome $y$, $loss(f(c_i),y)$, (2) a proximity factor, which is given by a distance function, and (3) a diversity factor $dpp\_diversity(c_1, ..., c_k)$. 
\begin{equation}
  \begin{split}
    C(x) = \underset{c_1,...,c_k}{\operatorname{arg min}} \frac{1}{k}\sum_{i=1}^{k} yloss(f(c_i),y) + \frac{\lambda_1}{k}\sum_{i=1}^{k} dist(c_i,x) \\
    - \lambda_2 \ ddp\_ diversity(c_1,...,c_k)~~~~~~~~~~~~~~~~~
    \end{split}
    \label{eq:dice}
\end{equation}

The distance function used in DiCE is the $L_1$-norm normalized by the inverse of the median absolute deviation of feature $j$ over the dataset. For optimisation, DiCE Gradient descent is used to minimize Equation~\ref{eq:dice}.

\subsection{Challenges of Counterfactuals in PPA and Event Logs} 

Contrary to traditional datasets, event logs are multi-dimensional data that turn the process of generating counterfactuals especially challenging due to several reasons:
\begin{enumerate}
    \item \textbf{Order matters.} One important property of event logs is the time sequence of events. When counterfactual algorithms attempt to find minimum changes in features that promote the desired outcome, these changes need to provide meaningful traces according to the process model domain knowledge.
    
    \item \textbf{Process domain knowledge should not be ignored.} Counterfactual algorithms for predictive process analytics need to consider the underlying processes that generated the event knowledge. Otherwise, when looking at minimum feature changes in traces, these may lead to counterfactual traces that are unrealistic according to the process model domain knowledge.
    
    \item \textbf{Traces with Lots of Activities are Beyond Human Interpretability.} Counterfactual algorithms for predictive process analytics need to consider the underlying processes that generated the event knowledge. Otherwise, when looking at minimum feature changes in traces, these may lead to counterfactual traces that are unrealistic according to the process model domain knowledge.
\end{enumerate}

\subsection{The Challenge of Generating Counterfactuals}
\label{categorical_issue}

\begin{figure*}[!h]
    \centering
    \includegraphics[scale=0.25]{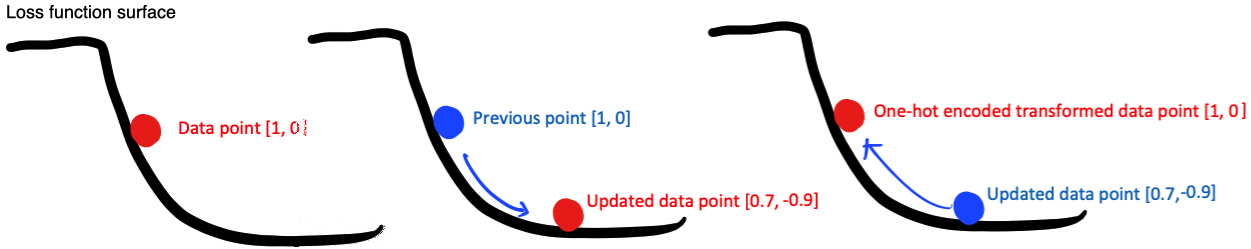}
    \caption{Categorical loss issue example. The nature of categorical variables can make the process of optimising and minimising a loss function challenging. At each learning step, gradient descent may not be able to find a local minimum of a function.}
    \label{fig:categorial_loss_issue}
\end{figure*}

Counterfactual algorithms such as DiCE consist of an optimization problem that minimizes the distance between the query instance and the counterfactual candidate. Although this is highly effective when the dataset's variables are real numbers, the optimization problem becomes challenging for categorical variables. Figure \ref{fig:categorial_loss_issue} provides an example of this challenge. Consider a categorical feature \textit{city}, which contains two different categories [Brisbane, Sydney]. The one-hot encoding representations of "Brisbane" and "Sydney" are given by the vectors [1, 0] and [0,1], respectively. In order to find a counterfactual, DiCE needs to locate a candidate vector whose distance is the minimum to the vector that represents Brisbane. To achieve this, DiCE uses a well-known optimizer widely used in neural networks called \textit{gradient descent}. At each iteration of the minimization process, gradient descent will change the input vectors into real numbers. For instance, the Brisbane one-hot encoded vector would become [0.7, -0.9], no longer one-hot encoded. To correct this, DiCE computes the maximum of each feature to the updated vector would become [1, 0], which is the same as the original input. To summarize, the transformation process of categorical variables can cause an infinite loop during the optimization process, and the gradient descent algorithm may not find a local minimum of the loss function, resulting in no counterfactuals are found. 

This problem is highly relevant in predictive process analytics since both event log activities, and resources correspond to categorical variables. Current explainable counterfactual algorithms in the literature cannot deal with event logs due to the problem of the categorical variables.

\section{DiCE4EL: Diverse Counterfactual Explanations for Event Logs}

We propose an extension of the DiCE algorithm to generate counterfactual explanations for process prediction using event logs, and refer to this new algorithm as DiCE4EL.

Given a prediction from a machine learning algorithm (either next activity prediction or outcome prediction), the proposed DiCE4EL algorithm starts by retrieving this prediction from the learning model and computes an opposite class to that prediction, which corresponds to the counterfactual class that we wish to get an explanation. For instance, if the learning model predicts that a loan application was `\textit{not approved}', the opposite counterfactual class will be `\textit{approved}'.

DiCE4EL can deal with both  the process context knowledge and categorical variables by minimizing a loss function that consists in a linear combination of four different sub-loss functions:
\begin{itemize}
\item \textbf{Class Loss}, which is the difference between the output of predicting model, $y$, and the opposite class, $\bar{y}$. We used the hinge loss is used to measure this difference;
\item \textbf{Distance Loss}, which minimizes the distance between an input query and a generated counterfactual candidate. In this loss, we apply gradient descent to minimize the distance function. In this step, categorical variables are treated as numeric variables to allow for convergence of gradient descent. We used the Euclidean distance (L2 norm) to compute the similarity between the vectors;
\item \textbf{Category Loss}, which corrects the categorical variables computed in the distance loss function by constraining the variables to the sum of the input query;
\item \textbf{Scenario Loss}, which ensures that the generated counterfactual exists in the process domain knowledge. The scenario validity is achieved by using the training set of the event log as the background knowledge of all known plausible traces (sequences of events of a case in an event log). This approach is inspired by Case-Based Reasoning approaches for counterfactual generation, where a database of valid counterfactuals is stored in the system~\cite{Keane20}. Our algorithm checks how the generated counterfactual is similar to the counterfactuals in the training set. If the output of this loss function is 1, then it means that the scenario model has a high chance to represent a valid counterfactual scenario. 
\end{itemize}
 
\begin{figure}[h!]
    \centering
    \includegraphics[scale=0.42]{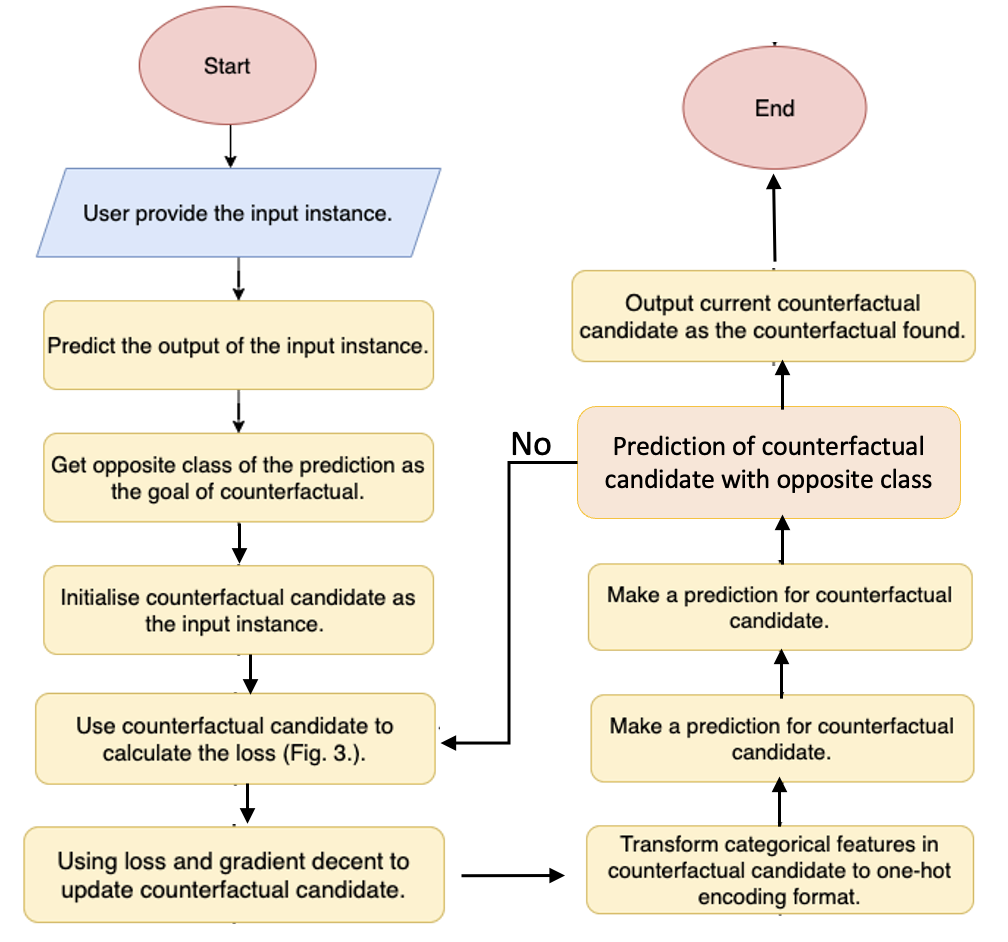}
    \caption{Design flowchart of DiCE4EL Algorithm}
    \label{fig:dice_flowchart}
\end{figure}

\begin{figure}[h!]
    \centering
    \includegraphics[scale=0.28]{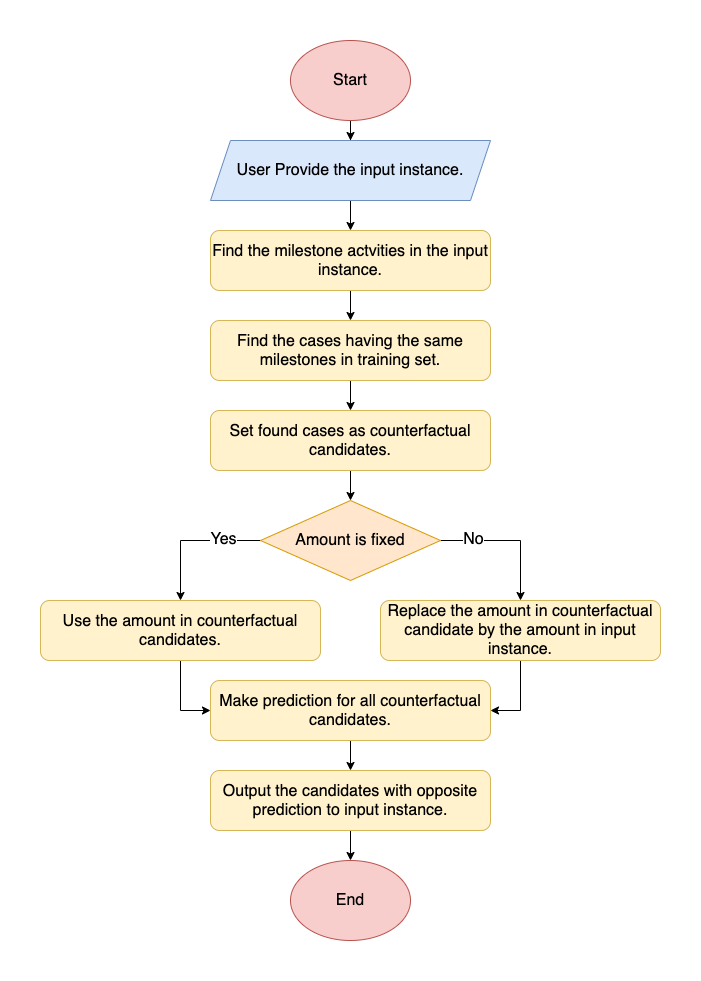}
    \caption{Counterfactual search in a training set according to DiCE4EL}
    \label{fig:cf_searching_flowchart}
\end{figure}

The proposed algorithm iteratively repeats the calculation and minimization of the above loss functions until the predictive model predicts the opposite class $\bar{y}$. Figures~\ref{fig:dice_flowchart} and~\ref{fig:cf_searching_flowchart} present a high level flowchart illustrating how DiCE4EL algorithm works. For more  details about the algorithm, the reader can refer to our open-source repository, which can be found in~\cite{DICE_RICH}.

\section{Approach}
\label{sec:approach}

The proposed approach is divided into three steps: 
i) identifying the milestone activities given an input event log; 
ii) predicting the next activity of a (process execution) trace; and 
iii) generating milestone-aware explainable counterfactuals using DiCE4EL. 
This paper uses milestone events of a loan application process for predicting the next activity and generates the respective explanations through counterfactuals using DiCE4EL. We create a neural network architecture that receives a sequence of activities, the related resources, and the loan amount of the application. Our predictive model architecture considers both dynamic (activities and resources) and static (amount) features. We then generate and analyse counterfactual explanations for process prediction towards each of the milestones.

\subsection{Dataset}

The event log that we use in this work is taken from a bank in the Netherlands and corresponds to a loan application, where customers request a certain amount of money. This dataset has been provided for the BPI Challenge in 2012 and is publicly available\footnote{http://www.win.tue.nl/bpi/doku.php?id=2012:challenge}. 

The loan application starts with a webpage from where a customer selects a certain amount of money and then submits his request. Then, the application performs some automatic tasks and checks if an application is eligible. If eligible, the customer is sent an offer by mail (or by phone). After this offer is received, it will be evaluated. In case of any missing information, the offer goes back to the client and is re-evaluated until all the required information is gathered. A final evaluation is then performed, and the application is approved.

\begin{figure}[!b]
    \centering
    \includegraphics[scale = 0.3]{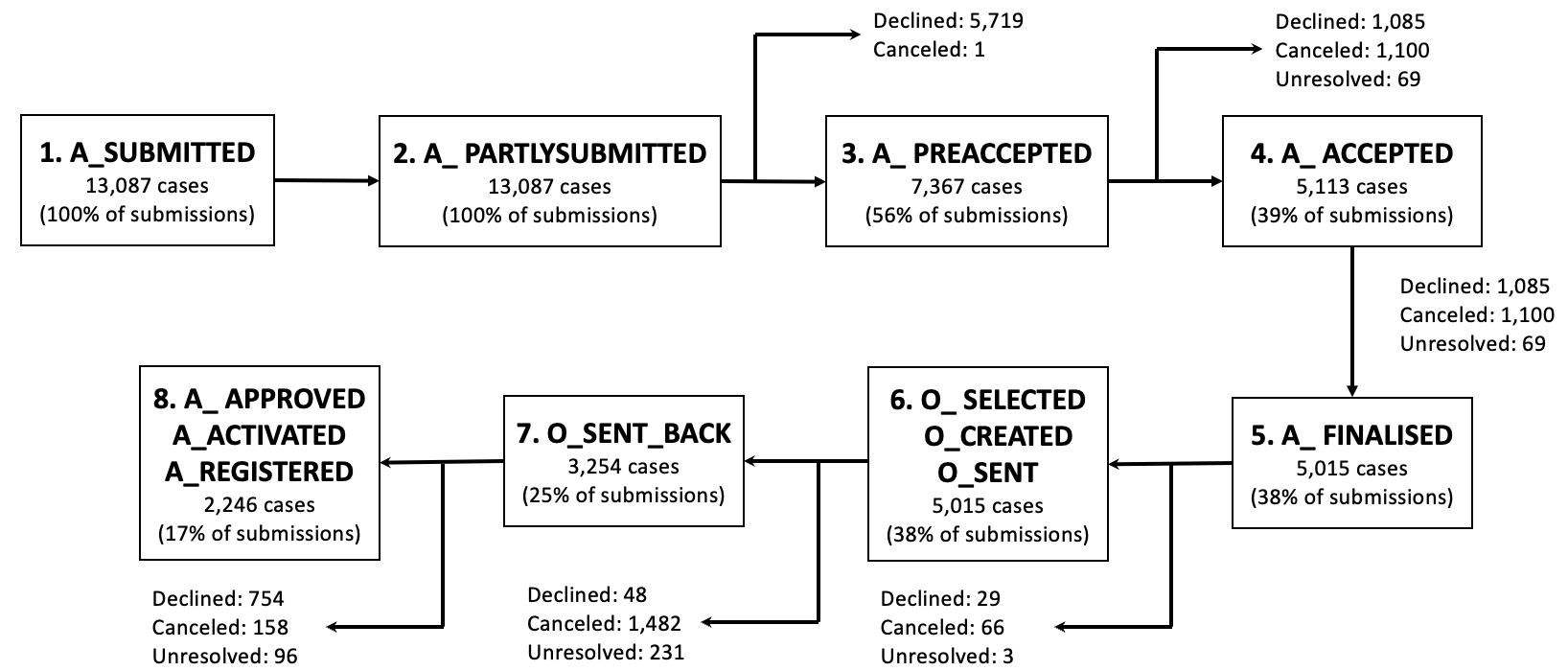}
    \caption{Milestones of loan application process captured in BPIC2012 as identified in \cite{Bautista13}}
    \label{fig:milestones}
\end{figure}

There are several activities that can be considered milestones in this event log, as previously identified in the winning report of BPI Challenge 2012~\cite{Bautista13} and are illustrated in Figure~\ref{fig:milestones}.


\subsection{Predicting the next activity: Neural Network Architecture}

The predictive model architecture that we use in this paper is inspired by the specialized architecture initially proposed in the work of~\cite{Camargo2019}. The main difference of this architecture is the separation of the dynamic features (such as activity or resource) from the static features (such as AMOUNT). The dynamic features are processed by a Long Short Term Memory (LSTM) neural network, while the static features are processed by a fully connected dense deep neural network. We concatenate these different features and compute the next activity of a given loan application. Details about the neural network architecture can be found in our open source repository \cite{DICE_RICH}.

Our model achieved an overall accuracy of $82\%$, precision $0.8$, recall $82\%$ and F1-score $79\%$. Table \ref{tab:training_parameters} presents the parameters that were used to configure the proposed neural network.


\begin{table}[h!]
    \centering
    \caption{training parameters}
    \begin{tabular}{|l|c|}
    \hline
    Parameters & Predictive model \\
    \hline
        Activity embedding dimension &  32  \\
        Resource embedding dimension & 128  \\
        LSTM hidden dimension & 64  \\
        Fully-connected layer dimension & 64  \\
        Dropout rate & 0.1 \\
        Epochs & 20 \\
        Batch size & 128  \\
        Optimizer & Adam  \\
        Learning rate & 0.005  \\
        Trainable Parameters & 169,306  \\
    \hline
    \end{tabular}
    \label{tab:training_parameters}
\end{table}

\subsection{Milestone-Aware Counterfactual Generation}

Given that there are no standardized evaluation protocols for counterfactual evaluation in XAI \cite{Carvalho19}, we perform a milestone-aware counterfactual generation process where we test different scenarios and analyzed the counterfactuals returned by DiCE4EL.

Given a set of milestones as indicated in Figure \ref{fig:milestones}, we test traces from the test set and use the proposed neural network to predict the next activity. If the predicted next activity is different from the (desired) milestone, we ask for an explanation in the form of a counterfactual:  \textit{what would I have had to change in my loan application in order to meet the milestone?} To answer this question, we apply the counterfactual generator that will indicate what features need to be changed in order to meet the desired milestone. To validate the quality of the generated counterfactual, we use the process domain knowledge to understand if the generated counterfactual corresponds to a legitimate scenario or an unrealistic one. As a more concrete example, given a sequence A\_SUBMITTED $\rightarrow$ A\_PARTLYSUBMITTED  $\rightarrow$ A\_PREACCEPTED $\rightarrow$ W\_Completeren aanvraag, and AMOUNT = \$500,000, the next activity predicted is A\_DECLINED. We then would like to understand what we need to change to have the loan application A\_ACCEPTED? Our counterfactual could say that for the loan application to be accepted, an AMOUNT = 10,000 would be required, suggesting that the client is requesting a very high loan amount in his/her current application.

\section{Evaluation}

This section presents the main results of our approach. It is important to note that there are no standardized evaluation protocols to assess the quality of generated counterfactuals. This constitutes an open research question in the field \cite{Carvalho19}. In this paper, we evaluate the quality of the generated counterfactuals by analyzing the properties to generate good counterfactuals as presented in \cite{Moreira21infofusion, Keane20}: sparsity, plausibility, proximity, diversity, and categorical. \textbf{Sparsity} corresponds to the average number of feature changes in the generated counterfactuals. The minimum number of changes, the more interpretable the counterfactual becomes. \textbf{Plausibility} emphasizes that the generated counterfactuals should be legitimate, and the search process should ensure logically reasonable results.
\textbf{Proximity} corresponds to the distance between the generated counterfactual and the input query. The smaller the distance, the more interpretable the counterfactual is. \textbf{Diversity} Corresponds to the capacity of generating different types of counterfactuals. \textbf{Categorical} means if the algorithm can deal with categorical variables (this is a fundamental property for counterfactual explanations in predictive process analytics).
 
 We also perform milestone-aware counterfactual analysis where we investigate, for specific inputs, what are the counterfactuals generated, and we assess if they make sense given the relevant process knowledge (e.g. the process map).
 
 \subsection{Evaluation of Counterfactual Properties}
 
Table~\ref{tab:comparion} shows the comparison between the proposed DiCE4EL and the original DiCE algorithm for two perspectives of process execution: \textit{activity} and \textit{resource} in the context of the BPIC2012 event log. DiCE cannot find counterfactuals in event logs, because of the multi-perspective nature of the event log data, and the representation of categorical variables. Consequently, the optimization process during gradient descent will fall in an endless loop. On the other hand, one can see that the proposed DiCE4EL algorithm can find counterfactuals. The counterfactuals generated constitute plausible explanations that are present in the process map. The algorithm is also able to find counterfactuals with the minimum feature change. We measured Sparsity using the Levenshtein and proximity using the Euclidean distance. On average, DiCE4EL found counterfactuals with 5 changes in the activities of a trace, and an Euclidean distance of 2.3. 
These results suggest that the closest counterfactual to the query has been found.  

\begin{table*}[]
\caption{Comparison between DiCE~\cite{Mothilal20} and DiCE4EL (this paper) in terms of different properties for the BPIC2012 event log. We measured sparsity using the average Levenshtein distance, and proximity using the L2 norm (Euclidean distance) between the returned counterfactuals and the query. }
\label{tab:comparion}\centering
\resizebox{1.5\columnwidth}{!} {
\begin{tabular}{lcc|c|c|c|c|c|}
\cline{4-8}
& \multicolumn{1}{l}{} & \multicolumn{1}{l|}{} & \multicolumn{5}{c|}{\cellcolor[HTML]{C0C0C0}{\color[HTML]{000000} \textbf{Properties}}}    \\ 
\hline
\rowcolor[HTML]{EFEFEF} 
\multicolumn{1}{|l|}{\cellcolor[HTML]{EFEFEF}\textbf{Algorithm}} & \multicolumn{1}{l|}{\cellcolor[HTML]{EFEFEF}\textbf{Code}} & \multicolumn{1}{l|}{\cellcolor[HTML]{EFEFEF}\textbf{Dimension}} & \multicolumn{1}{l|}{\cellcolor[HTML]{EFEFEF}\textbf{Proximity}} & \multicolumn{1}{l|}{\cellcolor[HTML]{EFEFEF}\textbf{Sparcity}} & \multicolumn{1}{l|}{\cellcolor[HTML]{EFEFEF}\textbf{Diversity}} & \multicolumn{1}{l|}{\cellcolor[HTML]{EFEFEF}\textbf{Plausbility}} & \multicolumn{1}{l|}{\cellcolor[HTML]{EFEFEF}\textbf{Categorical?}} \\ 
\hline
\multicolumn{1}{|l|}{}
& \multicolumn{1}{c|}{}  & Activity   & Not Found  & Not Found  & \greencheck  & \crosscheck & \crosscheck \\ 
\cline{3-8} 
\multicolumn{1}{|l|}{\multirow{-2}{*}{DiCE}}   & \multicolumn{1}{c|}{\multirow{-2}{*}{ \cite{Mothilal20} }}                   & Resource  & Not Found & Not Found & \greencheck  & \crosscheck    & \crosscheck   \\ 
\hline
\multicolumn{1}{|l|}{}  
& \multicolumn{1}{c|}{} & Activity  & 2.3531  & 5.4492  & \greencheck  & \greencheck & \greencheck \\ 
\cline{3-8} 
\multicolumn{1}{|l|}{\multirow{-2}{*}{DiCE4EL}}                  & \multicolumn{1}{c|}{\multirow{-2}{*}{ \cite{DICE_RICH}}} & Resource   & 2.6564   & 8.2947   & \greencheck   & \greencheck    & \greencheck                                                         \\ \hline
\end{tabular}
}
\end{table*}

\subsection{Milestone-Aware Counterfactual Analysis}

We performed different analyses where we conducted a milestone-aware counterfactual generation. The counterfactual was found by changing different features for the input query: keeping AMOUNT fixed \textit{v.s.} changing AMOUNT (together with resource and activity). We define the input as a query consisting of i) a sequence of events, where each event is denoted as a tuple of Activity, Resource, and Amount; ii) the current prediction; and iii) the (desired) milestone. In the following, we present examples of counterfactuals that were generated. 
For more details about our experiments, 
the reader can refer to our publicly available notebooks \cite{DICE_RICH}.\\

\noindent
\textit{1) Generating Counterfactuals for A\_ACCEPTED Milestone} \\
\noindent
Input Query:
\begin{itemize}
    \item[]
    (A\_SUBMITTED, 112, \$15,500),\\ 
   (A\_PARTLYSUBMITTED, 112, \$15,500),\\ 
    A\_PREACCEPTED, 112, \$15,500),\\ 
    (W\_Complete request, 11180, \$15,500),\\ 
    (W\_Complete request, 11201, \$15,500) 
\end{itemize}
\noindent   
Prediction: W\_Complete request \\
\noindent
Milestone: A\_ACCEPTED \\
\noindent
Counterfactual: \textit{What would I have had to change to have for the loan to be A\_ACCEPTED?}\\

 Table~\ref{tab:input1_accepted_fixed} shows the counterfactuals found by the DiCE4EL algorithm for A\_ACCEPTED milestone and with AMOUNT unchanged. And Table \ref{tab:input1_accepted} shows the counterfactuals found for A\_ACCEPTED milestone and with AMOUNT varying.\\
 
\begin{table}[h!]
\caption{Counterfactuals generated for INPUT 1. AMOUNT was fixed to \$15,500 as input query.}
 \resizebox{\columnwidth}{!} {
\begin{tabular}{|l|l|l|l|l|l|}
\hline
\multicolumn{2}{|l|}{\textbf{Counterfactual 1}} & \multicolumn{2}{l|}{\textbf{Counterfactual 2}} & \multicolumn{2}{l|}{\textbf{Counterfactual 3}} \\ \hline
\textbf{Activity}         & \textbf{Resource}       & \textbf{Activity}         & \textbf{Resource}      & \textbf{Activity}         & \textbf{Resource}      \\ \hline
A\_SUBMITTED            & 112   & A\_SUBMITTED          & 112       & A\_SUBMITTED          & 112                \\
A\_PARTLYSUBMITTED      & 112   & A\_PARTLYSUBMITTED    & 112       & A\_PARTLYSUBMITTED    & 112                \\
A\_PREACCEPTED          & 112   & A\_PREACCEPTED        & 10910     & A\_PREACCEPTED        & 10910              \\
A\_ACCEPTED             & 10931 & W\_Complete request   & 10912     & W\_Handling leads     & 10910              \\
---                     & ---   & A\_ACCEPTED           & 10912     & A\_ACCEPTED           & 10932              \\ 
\hline
\end{tabular}
}
\label{tab:input1_accepted_fixed}
\end{table}

\begin{table}[h!]
\caption{Counterfactuals generated for INPUT 1.}
 \resizebox{\columnwidth}{!} {
\begin{tabular}{|l|l|l|l|l|l|}
\hline
\multicolumn{2}{|l|}{\textbf{Counterfactual 1}} & \multicolumn{2}{l|}{\textbf{Counterfactual 2}} & \multicolumn{2}{l|}{\textbf{Counterfactual 3}} \\ \hline
\textbf{Activity}  & \textbf{Resource} & \textbf{Activity} & \textbf{Resource} & \textbf{Activity}  & \textbf{Resource}\\ 
\hline
A\_SUBMITTED            & 112       & A\_SUBMITTED          & 112       & A\_SUBMITTED          & 112       \\
A\_PARTLYSUBMITTED      & 112       & A\_PARTLYSUBMITTED    & 112       & A\_PARTLYSUBMITTED    & 112       \\
A\_PREACCEPTED          & 112       & A\_PREACCEPTED        & 112     & A\_PREACCEPTED        & 10629     \\
A\_ACCEPTED             & 10932     & W\_Complete request   & 11302     & W\_Handling leads     & 10629     \\
---                     & ---       & A\_ACCEPTED           & 10932     & A\_ACCEPTED           & 10932     \\ 
\hline
AMOUNT:                 & 8750      & AMOUNT:               & 25000     & AMOUNT:               & 14000  \\
\hline
\end{tabular}
}
\label{tab:input1_accepted}
\end{table} 

\noindent
\textit{2) Generating Counterfactuals for A\_FINALISED milestone}

\noindent
Input Query:
\begin{itemize}
    \item[]
    (A\_SUBMITTED, 112, \$15,500),\\ 
   (A\_PARTLYSUBMITTED, 112, \$15,500),\\ 
    A\_PREACCEPTED, 112, \$15,500),\\ 
    (A\_ACCEPTED, 10939, \$15,500),
\end{itemize}
\noindent   
Prediction: O\_SELECTED \\
\noindent
Milestone:  A\_FINALISED \\
\noindent
Counterfactual: \textit{What would I have had to change for the loan to be A\_FINALISED?}\\

Table \ref{tab:input2_finalised_fixed} shows the counterfactuals found by the DiCE4EL algorithm for A\_FINALISED milestone and with AMOUNT unchanged. Table \ref{tab:input2_finalised} shows the counterfactuals found for MILESTONE = A\_FINALISED and with AMOUNT varying. \\

\begin{table}[b!]
\caption{Counterfactuals generated for input query 2. AMOUNT was fixed to \$15,500 as input query.}
 \resizebox{\columnwidth}{!} {
\begin{tabular}{|l|l|l|l|l|l|}
\hline
\multicolumn{2}{|l|}{\textbf{Counterfactual 1}} & \multicolumn{2}{l|}{\textbf{Counterfactual 2}} & \multicolumn{2}{l|}{\textbf{Counterfactual 3}} \\ \hline
\textbf{Activity}         & \textbf{Resource}       & \textbf{Activity}         & \textbf{Resource}      & \textbf{Activity}         & \textbf{Resource}      \\ \hline
A\_SUBMITTED            & 112   & A\_SUBMITTED          & 112       & A\_SUBMITTED          & 112                \\
A\_PARTLYSUBMITTED      & 112   & A\_PARTLYSUBMITTED    & 112       & A\_PARTLYSUBMITTED    & 112                \\
A\_PREACCEPTED          & 112   & A\_PREACCEPTED        & 10910     & A\_PREACCEPTED        & 10939              \\
A\_ACCEPTED             & 10931 & W\_Complete request   & 10912     & W\_Handling leads     & 10939              \\
A\_FINALISED            & 10931 & A\_ACCEPTED           & 10932     & A\_ACCEPTED           & 11189              \\ 
---                     & ---   & A\_FINALISED          & 10932     & O\_SELECTED           & 11189             \\
---                     & ---   & ---                   & ---       & A\_FINALISED          & 11189             \\
\hline
\end{tabular}
}
\label{tab:input2_finalised_fixed}
\end{table}

\begin{table}[b!]
\caption{Counterfactuals generated for INPUT 2. }
 \resizebox{\columnwidth}{!} {
\begin{tabular}{|l|l|l|l|l|l|}
\hline
\multicolumn{2}{|l|}{\textbf{Counterfactual 1}} & \multicolumn{2}{l|}{\textbf{Counterfactual 2}} & \multicolumn{2}{l|}{\textbf{Counterfactual 3}} \\ \hline
\textbf{Activity}         & \textbf{Resource}       & \textbf{Activity}         & \textbf{Resource}      & \textbf{Activity}         & \textbf{Resource}      \\ \hline
A\_SUBMITTED            & 112       & A\_SUBMITTED          & 112       & A\_SUBMITTED        & 112             \\
A\_PARTLYSUBMITTED      & 112       & A\_PARTLYSUBMITTED    & 112       & A\_PARTLYSUBMITTED  & 112       \\
A\_PREACCEPTED          & 112       & A\_PREACCEPTED        & 112     & A\_PREACCEPTED        & 112     \\
A\_ACCEPTED             & 11319     & W\_Complete request   & 11179     &  A\_ACCEPTED        & 10982      \\
A\_FINALISED            & 11319     & A\_ACCEPTED           & 10912     &  O\_SELECTED        & 10982 \\
---                     & ---       & A\_FINALISED          & 10912     &  A\_FINALISED       & 10982  \\
\hline
\textbf{AMOUNT}:        & 20000     & AMOUNT:               &  30000         & AMOUNT:        & 8000 \\
\hline
\end{tabular}
}
\label{tab:input2_finalised}
\end{table}

\noindent
\textit{3) Generating Counterfactuals for A\_APPROVED milestone}

\noindent
Input Query:
\begin{itemize}
    \item[]
    (A\_SUBMITTED, 112, \$15,500),\\ 
    (A\_PARTLYSUBMITTED, 112, \$15,500),\\ 
    A\_PREACCEPTED, 112, \$15,500),\\ 
    (A\_ACCEPTED, 10138, \$15,500), \\
    (A\_FINALIZED, 10138, \$15,500), \\
    (O\_SELECTED, 10138, \$15,500), \\
    (O\_CREATED, 10138, \$15,500), \\
    (O\_SENT, 10138, \$15,500), \\
    (W\_Complete request, 10138, \$15,500), \\
    (O\_SENT\_BACK, 10138, \$15,500), \\
    (W\_Calling quote, 10138, \$15,500),
\end{itemize}
\noindent   
Prediction: W\_Validate request \\
\noindent
Milestone:  A\_APPROVED  \\
\noindent
Counterfactual: \textit{What would I have had to change for the loan to be A\_APPROVED?}\\

Table~\ref{tab:input3_approved_fixed} shows the counterfactuals found by the DiCE4EL algorithm for A\_APPROVED milestone and with AMOUNT unchanged. Table~\ref{tab:input3_approved} shows the counterfactuals found for A\_APPROVED milestone and with AMOUNT varying.

\begin{table}[b!]
\caption{Counterfactuals generated for Input 3. AMOUNT was fixed to \$15,500 as input query.}
 \resizebox{\columnwidth}{!} {
\begin{tabular}{|l|l|l|l|l|l|}
\hline
\multicolumn{2}{|l|}{\textbf{Counterfactual 1}} & \multicolumn{2}{l|}{\textbf{Counterfactual 2}} & \multicolumn{2}{l|}{\textbf{Counterfactual 3}} \\ \hline
\textbf{Activity}         & \textbf{Resource}       & \textbf{Activity}         & \textbf{Resource}      & \textbf{Activity}         & \textbf{Resource}      \\ \hline
A\_SUBMITTED            & 112     & A\_SUBMITTED            & 112       & A\_SUBMITTED          & 112                \\
A\_PARTLYSUBMITTED      & 112     & A\_PARTLYSUBMITTED      & 112       & A\_PARTLYSUBMITTED    & 112                \\
A\_PREACCEPTED          & 112     & A\_PREACCEPTED          & 10910     & A\_PREACCEPTED        & 10939              \\
A\_ACCEPTED             & 10138   & W\_Complete request     & 10912     & W\_Handling leads     & 10913              \\
A\_FINALISED            & 10138   & A\_ACCEPTED             & 10932     & A\_ACCEPTED           & 10913              \\ 
O\_SELECTED             & 10138   & O\_SELECTED             & 10932     & O\_SELECTED           & 10913             \\
O\_CREATED              & 10138   & A\_FINALIZED            & ---       & A\_FINALISED          & 10913             \\
O\_SENT                 & 10138   & O\_CREATED              & ---       & O\_CREATED            & 10913               \\
W\_Complete request     & 10138   & O\_SENT                 & ---       & O\_SENT               & 10913               \\
O\_SENT\_BACK            & 10138   & W\_Complete request     & ---       & W\_Complete request   & 10913               \\
W\_Validate request     & 10138   & O\_SENT\_BACK           & ---       & O\_SENT\_BACK         & 10789               \\
A\_REGISTERED           & 10138   & W\_Complete request     & ---       & W\_Validate request   & 10789              \\
A\_APPROVED             & 10138   & W\_Validate request     & ---       & A\_REGISTERED         & 10789           \\
---                     & ---     & A\_ACTIVATED            & ---       & A\_APPROVED           & 10789             \\
\hline
\end{tabular}
}
\label{tab:input3_approved_fixed}
\end{table}

\begin{table}[b!]
\caption{Counterfactuals generated for Input 3. AMOUNT was fixed to \$15,500 as input query.}
 \resizebox{\columnwidth}{!} {
\begin{tabular}{|l|l|l|l|l|l|}
\hline
\multicolumn{2}{|l|}{\textbf{Counterfactual 1}} & \multicolumn{2}{l|}{\textbf{Counterfactual 2}} & \multicolumn{2}{l|}{\textbf{Counterfactual 3}} \\ \hline
\textbf{Activity}         & \textbf{Resource}       & \textbf{Activity}         & \textbf{Resource}      & \textbf{Activity}         & \textbf{Resource}      \\ \hline
A\_SUBMITTED            & 112     & A\_SUBMITTED            & 112       & A\_SUBMITTED          & 112                \\
A\_PARTLYSUBMITTED      & 112     & A\_PARTLYSUBMITTED      & 112       & A\_PARTLYSUBMITTED    & 112                \\
A\_PREACCEPTED          & 112     & A\_PREACCEPTED          & 112       & A\_PREACCEPTED        & 11203              \\
A\_ACCEPTED             & 10138   & W\_Complete request     & 11203     & W\_Handling leads     & 11203              \\
A\_FINALISED            & 10138   & A\_ACCEPTED             & 11203     & A\_ACCEPTED           & 10861              \\ 
O\_SELECTED             & 10138   & O\_SELECTED             & 11203     & O\_SELECTED           & 10861             \\
O\_CREATED              & 10138   & A\_FINALIZED            & 11203     & A\_FINALISED          & 10861             \\
O\_SENT                 & 10138   & O\_CREATED              & 11203     & O\_CREATED            & 10861               \\
W\_Complete request     & 10138   & O\_SENT                 & 11203     & O\_SENT               & 10861               \\
O\_SENT\_BACK            & 10138   & W\_Complete request     & 11203     & W\_Complete request   & 10861               \\
W\_Validate request     & 10138   & O\_SENT\_BACK           & 11049     & W\_Validate request   & 10789               \\
A\_APPROVED             & 10138   & W\_Validate request     & 11049     & O\_SENT\_BACK         & 10861              \\
---                     & --      & O\_ACCEPTED             & 10138     & W\_Validate request   & 10789           \\
---                     & ---     & A\_ACTIVATED            & 10138     & A\_REGISTERED         & 10789             \\
---                     & ---     & A\_APPROVED             & 10138     & A\_APPROVED           & 10138           \\
\hline
AMOUNT                  & 5000    & AMOUNT                  & 10000     & AMOUNT                & 20000                  \\
\hline
\end{tabular} 
} 
\label{tab:input3_approved}
\end{table}

\subsection{Discussion}

There are no standard evaluation protocols in the literature to determine the effectiveness of a counterfactual approach~\cite{Carvalho19}, let alone in the context of predictive process analytics. To address this problem, we used DISCO\footnote{\url{https://fluxicon.com/disco/}} to extract the process map and corresponding process context knowledge to assess the validity and meaningfulness of the generated counterfactuals. We concluded that all counterfactuals generated by the proposed algorithm are valid and correspond the shortest possible path from the input query to the desired outcome (in other words, the trace with the smallest amount of changes from the initial query).

\begin{figure}[h!]
    \centering
    \includegraphics[scale=0.3]{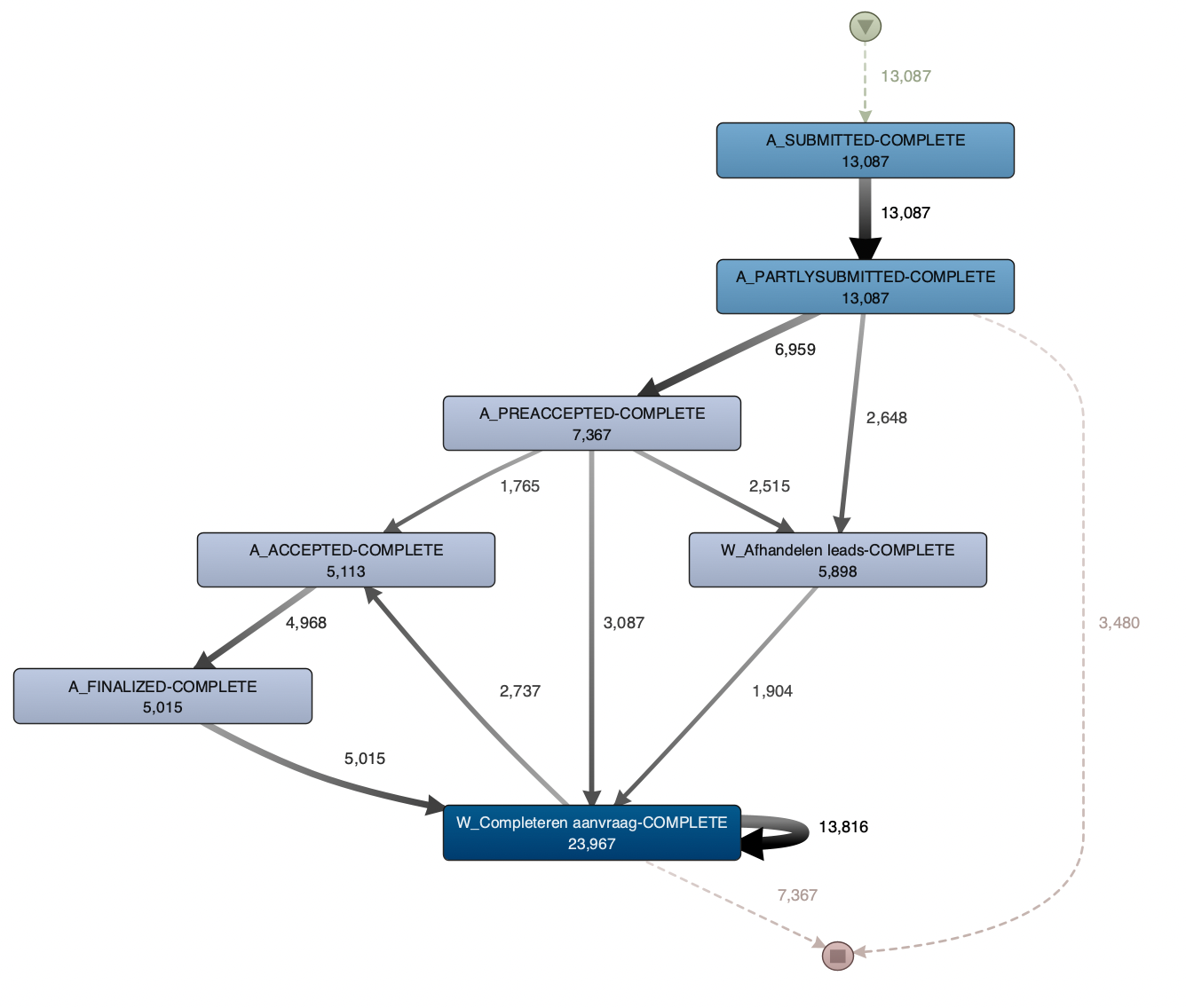}
    \caption{Fraction of process map extracted from DISCO, involving milestones A\_ACCEPTED and A\_FINALISED}
    \label{fig:disco}
\end{figure}

For input 1, we applied the proposed next activity classifier and obtained the prediction W\_Complete request, which suggests that this trace is in a loop. For business process analysis, it would be interesting to understand \textit{what would have needed to be changed in order to break this loop and move towards A\_ACCEPTED milestone}? The proposed counterfactual generator found several counterfactual scenarios to address this question. Table~\ref{tab:input1_accepted_fixed} presents all the counterfactuals found by considering the same AMOUNT of money requested in the original loan application process, and Table \ref{tab:input1_accepted} returns counterfactual scenarios where just changing the requested AMOUNT would lead to successfully reaching the milestone. For instance, according to the proposed algorithm, one potential change that could have been applied in the process execution to break the \textit{W\_Complete request} cycle would be to either: (1) use the resource $10931$ after the A\_PREACCEPTED milestone; (2) change the resource of A\_PREACCEPTED to $10912$, and process the activity W\_Complete request with that same resource; or (3) change the resource of A\_PREACCEPTED to $10910$, and change the activity W\_Complete request to W\_Handling leads with resource $10910$. All these scenarios are valid as the reader can verify in the portion of the process extracted from DISCO \ref{fig:disco}. Another interesting analysis is that if the client had requested the smaller AMOUNT = 8750, the process would transition from A\_PREACCEPTED $\rightarrow$ A\_ACCEPTED. The generated counterfactuals (Table \ref{tab:input1_accepted}) suggest that for high AMOUNTS, the process will proceed towards W\_Complete request, or it would proceed towards the activity W\_Handling leads. A similar analysis can be made for the experiments for the A\_FINALISED and A\_APPROVED milestones. A similar analysis can be done for input queries 2 and 3. For more detailed information, the reader can check our publicly available notebook that details all the steps that were taken in this research \cite{DICE_RICH}.
The counterfactuals generated and the analysis conducted in this work can equip business analysts with insights that can help them improve their processes, target potential unexpected outcomes, or even predict time-consuming processes and prepare strategies to mitigate the issues. Counterfactual approaches contribute to explainability by adding justifications of why the predictive model makes such predictions. These approaches are highly important to derive prescriptive business analysis, deviation analysis, or even model inspection.

\section{Conclusion}

This work analyzed three different queries, predicted their outcomes, and generated the respective counterfactuals with the proposed DiCE4EL algorithm. Counterfactual algorithms proposed in the literature are not suitable for predictive process analytics. They cannot generate realistic counterfactuals, majorly due to the multi-dimensional and sequential nature of the event logs (most algorithms require the application of permutations in the data that can lead to sequences of counterfactuals that are unrealistic), and due to optimization problems with categorical variables.

Our algorithm incorporates a \textit{valid scenario} function that searches for valid counterfactuals in the training set and uses that information in the loss function. The goal is to find the valid traces that require the minimum number of activity changes or amount changes. Our results indicate that our algorithm generates faithful and meaningful counterfactuals in accordance with the process domain knowledge. Unfortunately, there are no standard evaluation protocols to determine the effectiveness of our approach (how to assess what makes a counterfactual a good counterfactual). For future work, we are interested in defining a standard quantitative evaluation protocol that can be used in the context of predictive process analytics. As another direction of future work, we are interested in extending the proposed analysis using other types of event logs and also explore other well-known counterfactuals in XAI literature for predictive process analytics.

\section{Acknowledgements}

This work was supported by  Centre for Data Science First Byte Funding Program at Queensland University of Technology (QUT) and by QUT's Women in Research Grant Scheme.

\bibliographystyle{IEEEtran}

\begin{thebibliography}{10}
\providecommand{\url}[1]{#1}
\csname url@samestyle\endcsname
\providecommand{\newblock}{\relax}
\providecommand{\bibinfo}[2]{#2}
\providecommand{\BIBentrySTDinterwordspacing}{\spaceskip=0pt\relax}
\providecommand{\BIBentryALTinterwordstretchfactor}{4}
\providecommand{\BIBentryALTinterwordspacing}{\spaceskip=\fontdimen2\font plus
\BIBentryALTinterwordstretchfactor\fontdimen3\font minus
  \fontdimen4\font\relax}
\providecommand{\BIBforeignlanguage}[2]{{%
\expandafter\ifx\csname l@#1\endcsname\relax
\typeout{** WARNING: IEEEtran.bst: No hyphenation pattern has been}%
\typeout{** loaded for the language `#1'. Using the pattern for}%
\typeout{** the default language instead.}%
\else
\language=\csname l@#1\endcsname
\fi
#2}}
\providecommand{\BIBdecl}{\relax}
\BIBdecl

\bibitem{Teinemaa19}
I.~Teinemaa, M.~Dumas, M.~L. Rosa, and F.~M. Maggi, ``{Outcome-Oriented
  Predictive Process Monitoring: Review and Benchmark},'' \emph{ACM
  Transactions on Knowledge Discovery from Data}, vol.~13, 2019.

\bibitem{Verenich19}
I.~Verenich, M.~Dumas, M.~L. Rosa, F.~M. Maggi, and I.~Teinemaa, ``{Survey and
  Cross-Benchmark Comparison of Remaining Time Prediction Methods in Business
  Process Monitoring},'' \emph{ACM Transactions on Intelligent Systems and
  Technology}, vol.~10, no.~4, 2019.

\bibitem{neu2021systematic}
D.~A. Neu, J.~Lahann, and P.~Fettke, ``A systematic literature review on
  state-of-the-art deep learning methods for process prediction,''
  \emph{Artificial Intelligence Review}, pp. 1--27, 2021.

\bibitem{Lipton2018}
Z.~C. Lipton, ``The mythos of model interpretability,'' \emph{Communications
  {ACM}}, vol.~61, no.~10, pp. 36--43, 2018.

\bibitem{Doran2017}
D.~Doran, S.~Schulz, and T.~R. Besold, ``What does explainable ai really mean?
  a new conceptualization of perspectives,'' in \emph{Proceedings of the First
  International Workshop on Comprehensibility and Explanation in AI and ML 2017
  co-located with 16th International Conference of the Italian Association for
  Artificial Intelligence}, 2017.

\bibitem{serrano2019attention}
S.~Serrano and N.~A. Smith, ``Is attention interpretable?'' in \emph{Proc. of
  the 57th Conference of the Association for Computational Linguistics,
  {ACL}}.\hskip 1em plus 0.5em minus 0.4em\relax Association for Computational
  Linguistics, 2019, pp. 2931--2951.

\bibitem{Moreira20medical}
W.~Tan, P.~Tiwari, H.~M. Pandey, C.~Moreira, and A.~K. Jaiswal, ``Multi-modal
  medical image fusion algorithm in the era of big data,'' \emph{Neural
  Computing and Applications}, 2020.

\bibitem{EVERMANN2017129}
J.~Evermann, J.-R. Rehse, and P.~Fettke, ``{Predicting Process Behaviour Using
  Deep Learning},'' \emph{Decision Support Systems}, vol. 100, pp. 129--140,
  2017, smart Business Process Management.

\bibitem{Ribeiro16}
M.~T. Ribeiro, S.~Singh, and C.~Guestrin, ``"{W}hy {S}hould {I} {T}rust
  {Y}ou?": Explaining the predictions of any classifier,'' in \emph{Proceedings
  of the 22nd ACM SIGKDD International Conference on Knowledge Discovery and
  Data Mining}, 2016, pp. 1135--1144.

\bibitem{Lundberg17}
S.~Lundberg and S.-I. Lee, ``A unified approach to interpreting model
  predictions,'' in \emph{Proceedings of the 31st Annual Conference on Neural
  Information Processing Systems (NIPS)}, 2017, pp. 4765--4774.

\bibitem{Moreira21dss}
C.~Moreira, Y.-L. Chou, M.~Velmurugan, C.~Ouyang, R.~Sindhgatta, and P.~Bruza,
  ``Linda-bn: An interpretable probabilistic approach for demystifying
  black-box predictive models,'' \emph{Decision Support Systems}, vol. (in
  press), p. 113561, 2021.

\bibitem{Byrne1997}
R.~Byrne, ``Cognitive processes in counterfactual thinking about what might
  have been,'' \emph{The psychology of learning and motivation: Advances in
  research and theory}, vol.~37, pp. 105--154, 1997.

\bibitem{Poyiadzi20}
R.~Poyiadzi, K.~Sokol, R.~Santos-Rodriguez, T.~De~Bie, and P.~Flach, ``Face:
  Feasible and actionable counterfactual explanations,'' in \emph{Proceedings
  of the AAAI/ACM Conference on ai, ethics, and society}, 2020, pp. 344--350.

\bibitem{Mothilal20}
R.~Mothilal, A.~Sharma, and C.~Tan, ``Explaining machine learning classifiers
  through diverse counterfactual explanations,'' in \emph{Proceedings of the
  2020 Conference on fairness, accountability, and transparency}, 2020, pp.
  607--617.

\bibitem{Moreira21infofusion}
Y.-L. Chou, C.~Moreira, P.~Bruza, C.~Ouyang, and J.~Jorge, ``Counterfactuals
  and causability in explainable artificial intelligence: Theory, algorithms,
  and applications,'' 2021.

\bibitem{Verma20}
S.~Verma, J.~Dickerson, and K.~Hines, ``Counterfactual explanations for machine
  learning: A review,'' \emph{arxiv: 2010.10596}, 2020.

\bibitem{Keane20}
M.~T. Keane and B.~Smyth, ``Good counterfactuals and where to find them: A
  case-based technique for generating counterfactuals for explainable ai
  (xai),'' \emph{arxiv: 2005.13997}, 2020.

\bibitem{DICE_RICH}
\BIBentryALTinterwordspacing
``Dice4el,'' 2021. [Online]. Available:
  \url{https://github.com/ChihchengHsieh/EventLogDiCE}
\BIBentrySTDinterwordspacing

\bibitem{Bautista13}
A.~D. Bautista, L.~Wangikar, and S.~M.~K. Akbar, ``Process mining-driven
  optimization of a consumer loan approvals process,'' in \emph{Business
  Process Management Workshops}, 2013, pp. 219--220.

\bibitem{Camargo2019}
M.~Camargo, M.~Dumas, and O.~G. Rojas, ``Learning accurate {LSTM} models of
  business processes,'' in \emph{Proceedings of the 17th International
  Conference, {BPM}}, 2019, pp. 286--302.

\bibitem{Carvalho19}
D.~V. Carvalho, E.~M. Pereira, and J.~S. Cardoso, ``Machine learning
  interpretability: A survey on methods and metrics,'' \emph{Electronics},
  vol.~8, p. 832, 2019.

\end{thebibliography}


\end{document}